\documentclass[10pt,twocolumn,letterpaper]{article}

\usepackage{iccv}              

\definecolor{iccvblue}{rgb}{0.21,0.49,0.74}
\usepackage[pagebackref,breaklinks,colorlinks,allcolors=iccvblue]{hyperref}
\usepackage{algorithm}
\usepackage{algorithmic}
\usepackage{multicol}
\usepackage{multirow}
\usepackage[normalem]{ulem}
\useunder{\uline}{\ul}{}
\usepackage{graphicx}
\usepackage{multirow}
\usepackage[normalem]{ulem}

\usepackage{multicol}
\usepackage{subcaption}
\usepackage{tikz}
\usepackage{amsmath}
\usepackage{arydshln}
\def\paperID{10045} 
\def\confName{ICCV}
\def\confYear{2025}

\title{Object Affordance Recognition and Grounding via Multi-scale Cross-modal Representation Learning}

\author{
Xinhang Wan$^1$\thanks{Both authors contributed equally to this research.}\thanks{Work done when Xinhang Wan was a visiting student at ShanghaiTech University.}, 
Dongqiang Gou$^{2}$\footnotemark[1], 
Xinwang Liu$^{1}$, 
En Zhu$^{1}$, 
Xuming He$^2$\thanks{Corresponding author.}
\\
$^1$National University of Defense Technology, Changsha, China
$^2$ShanghaiTech University, Shanghai, China 
\\
{\tt\small \{wanxinhang, xinwangliu, enzhu\}@nudt.edu.cn}, 
{\tt\small \{goudq2023, hexm\}@shanghaitech.edu.cn}
}
\begin{document}
\maketitle
\begin{abstract}
A core problem of Embodied AI is to learn object manipulation from observation, as humans do. To achieve this, it is important to localize 3D object affordance areas through observation such as images (3D affordance grounding) and understand their functionalities (affordance classification). 
\textcolor{black}{Previous attempts usually tackle these two tasks separately, leading to inconsistent predictions due to lacking proper modeling of their dependency. 
}
In addition, these methods typically 
only ground the incomplete affordance areas depicted in images, failing to predict the full potential affordance areas, and 
operate at a fixed scale, 
resulting in difficulty in coping with affordances significantly 
varying in scale w.r.t the whole object.
To address these issues, we propose a novel approach that learns an affordance-aware 3D representation and employs a stage-wise inference strategy leveraging the dependency between grounding and classification tasks. Specifically, we first develop a cross-modal 3D representation through efficient fusion and multi-scale geometric feature propagation, enabling inference of full potential affordance areas at a suitable regional scale. Moreover, we adopt a simple two-stage prediction mechanism, \textcolor{black}{effectively coupling grounding and classification for better affordance understanding}. 
Experiments demonstrate the effectiveness of our method, showing improved performance in both affordance grounding and classification.
\end{abstract}    
\section{Introduction}
\label{sec:intro}


\begin{figure}
    \centering
    \includegraphics[width=1\linewidth]{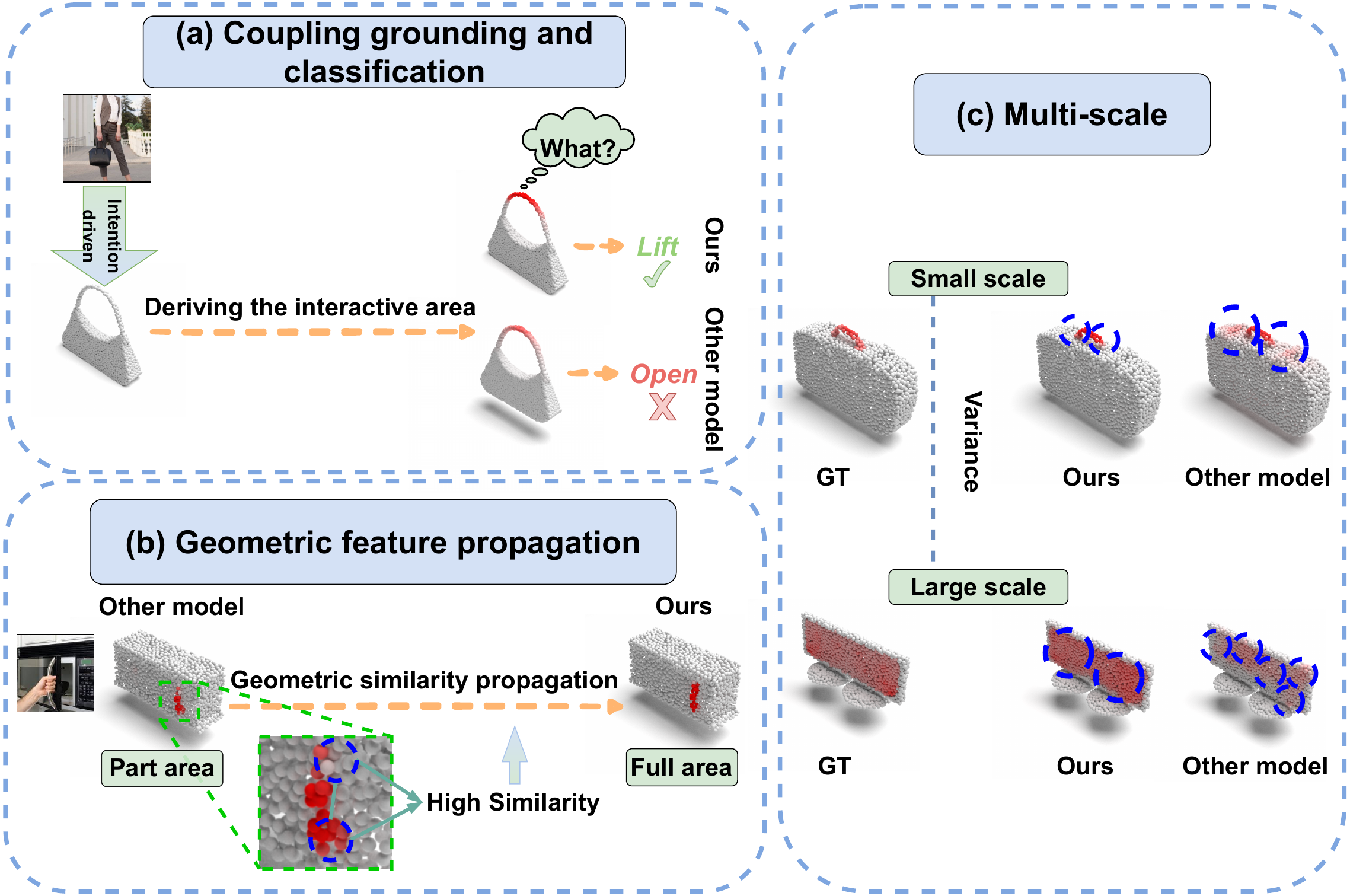}
    \caption{Motivation of Our Method: a) The cascaded model effectively couples grounding and classification; b) The inherent geometric similarity among different parts of an object allows for deriving the full potential interactive areas from a specific interaction; c) Learning affordance areas from a single scale is inefficient due to the wide span of scales, while multi-scale geometric features facilitate effective learning of affordance areas across this range of scales.}
    \label{motivation_fra}
    \vspace{-1em}
\end{figure}

\textcolor{black}{As robotics and computer vision advance, the significance of embodied AI becomes increasingly evident \cite{ZHANG2021100224,XU2021100179,pelau2021makes}. A core problem of embodied AI is learning object manipulation from observation, like humans do \cite{toth2009oldowan,ko2016origins}. A fundamental prerequisite for object manipulation is the ability to localize the interactive areas of 3D objects through observation (3D affordance grounding) and to understand their functionalities (affordance classification), as these aspects define where and how an agent can effectively interact with an object \cite{7280587,Deng_2021_ICCV,Zhan_Yuan_Xiong_2024,qian2024affordancellm,Zhang_2023_ICCV}.}
In this paper, we focus on both grounding and classifying 3D object affordances from images, enabling robots to understand how to manipulate 3D objects under the guidance of images.

Early efforts often address those two problems \cite{chuang2018learning, liu2024pavlm} separately. For instance, \cite{zhai2022one,10.1145/3446370} focus on affordance classification but primarily at the object level, lacking a fine-grained understanding of affordances. 
In contrast, other approaches \cite{Yang_2021_ICCV,xu2024weaklysupervised3dvisualgrounding} align 2D images with 3D point clouds, using the aligned representation to guide grounding and locate affordance areas. Recently, several works have attempted to address the two tasks simultaneously. In particular, IAG \cite{10378483} employs a 2D-3D alignment network to align images with point clouds, subsequently using this alignment representation to ground and classify 3D affordances independently. However, such a design is unable to fully capture the dependency between two tasks, often leading to inconsistent predictions (See Fig.\ref{motivation_fra}.(a)). 
Moreover, for the grounding task, existing methods often suffer from two main limitations: 1) a tendency to ground the incomplete regions depicted from the image, failing to predict the full potential affordance area (See Fig.\ref{motivation_fra}.(b)). 2) generating grounding at a fixed scale, whereas different affordance regions can vary greatly in their scope relative to the object (See Fig.\ref{motivation_fra}.(c)).

To address these challenges, we propose a novel cross-modal fusion framework for 3D affordance grounding and classification. Our method learns an effective affordance-aware 3D representation and employs a stage-wise inference strategy that leverages the dependency between grounding and classification tasks.
Specifically, we develop a cross-modal affordance representation through efficient fusion and multi-scale geometric feature propagation. This approach facilitates the inference of the full scope of affordance areas, accommodating regions of varying sizes (See Fig.~\ref{motivation_fra}(b)\&(c)). Additionally, we adopt a simple cascaded scheme for the grounding and classification tasks, effectively coupling them and alleviating inconsistencies between their predictions.


In detail, given a single image depicting a person interacting with an object and the corresponding 3D point cloud, we first employ an image encoder to extract context-aware affordance features from the 2D image. Recognizing that scene context plays a varying role in affordance prediction, we design a novel gating network that selectively integrates scene information. Concurrently, we utilize a point cloud encoder to extract multi-scale features from the 3D point cloud. We then propose a simple yet efficient cross-modal fusion module, consisting of two multi-head cross-attention mechanisms, to integrate features from both modalities. To better cover the full potential affordance area beyond the incomplete affordance region reflected in the image (See Fig.\ref{motivation_fra}.(b)), we exploit the geometric information of the point cloud to propagate regional features across multiple scales, ensuring that geometrically similar regions share similar representations. Subsequently, we leverage a multi-scale selector to fuse these multi-scale features. Finally, we predict the grounding probabilistic mask and then fuse the global and masked local features to predict the affordance class.

We conduct comprehensive experiments and both the statistical and visual results show that the effectiveness of our method, and validate the effectiveness of geometric feature propagation and multi-scale feature extraction.
In summary, our contributions are as follows:
\begin{enumerate}
    \item We propose a novel cross-modal multiscale affordance representation that can better handle large variations in affordance regions and improve the coverage of affordance grounding via efficient fusion and propagation.
    \item We employ a simple yet effective cascaded strategy for coupling the affordance grounding and classification tasks, which mitigates the inconsistencies between their predictions.
    \item We comprehensively evaluate both grounding and classification performance on the standard benchmark. Extensive numerical and visual results demonstrate the effectiveness of our method.
\end{enumerate}



\section{Related Work}
\label{sec:formatting}
\subsection{Affordance leanring}
Existing works on affordance learning focuses on human-made objects, as these objects are designed to fulfill specific functional or interactive needs \cite{roy2016multi,8460902,LUDDECKE201992,7759429}. Early research originated in the image domain, with the primary task being to recognize interactive objects in images and classify their affordance types \cite{Chuang_2018_CVPR,mi2020intention,chen2023survey}. However, they are often limited to the object level, lacking fine-grained affordance understanding.

With the introduction of embodied AI, the fine-grained manipulation and labeling of 3D objects has become an area of growing interest \cite{NEURIPS2023_0e7e2af2,tabib2024lgafford,Delitzas_2024_CVPR}. 3D AffordanceNet \cite{deng20213d} introduced the first fine-grained labeled dataset for 3D objects, where each point in the point cloud is annotated with interaction probabilities. Inspired by this dataset, a series of researches have been proposed. For instance, Jact et al.~\cite{merullo2022pretraining}, combined natural language processing (NLP) with affordance learning, demonstrating that 3D information is more suitable for affordance learning than 2D data. Additionally, PartAfford \cite{xu2022partafford} segments 3D point clouds into affordance-specific components.

However, existing approaches often overlook a deeper understanding of the specific operations associated with affordances, typically treating affordance as a mere label. For instance, PartAfford can segment a chair into components labeled as "sit", but it does not provide an understanding of the specific actions that correspond to the "sit" affordance. Humans learn affordances by observing interactions, whereas 3D point clouds alone fail to capture interaction dynamics.

\subsection{Image-Point Cloud Cross-Modal Learning}
\begin{figure*}
    \centering
    \includegraphics[width=1\linewidth]{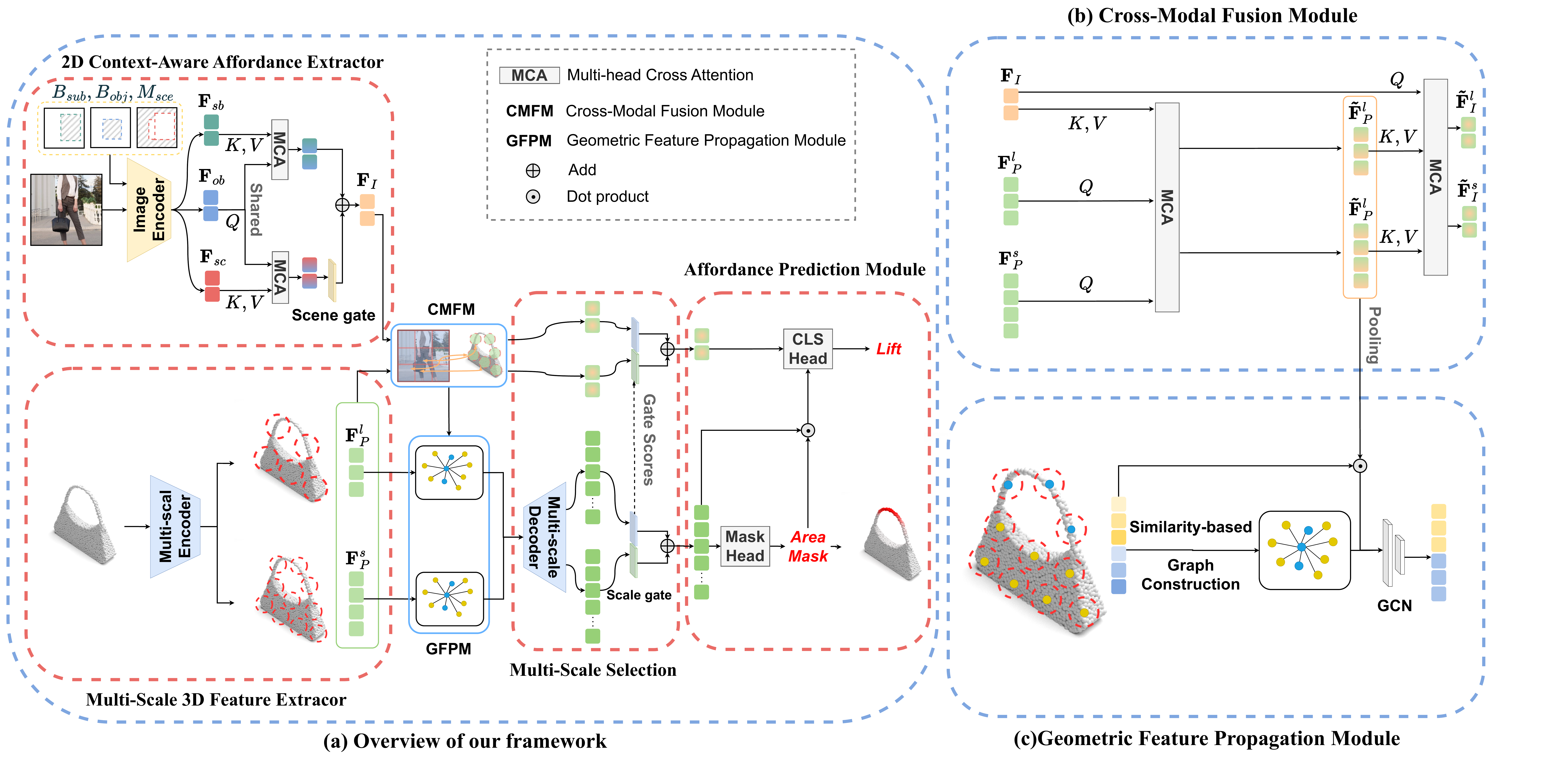}
    \caption{Overview of our framework. It identifies the affordance region and category in four steps: 1) Extract 2D context-aware affordance feature $\mathbf{F}_{I}$ and multi-scale 3D geometric features $\mathbf{F}_{P}^{l}$ and $\mathbf{F}_{P}^{s}$(Sec. \ref{feature_extra}); 2) Fuse the two modalities, obtain multiscale features $\mathbf{\tilde{F}}_P^l$, 
    $\mathbf{\tilde{F}}_P^s$, $\mathbf{\tilde{F}}_I^l$ and $\mathbf{\tilde{F}}_I^s$  (Sec. \ref{CMFM}); 3) Propogate the regional features at each scale and conduct scale selection to generate final representations $\tilde{\mathbf{R}}_{P}$ and $\tilde{\mathbf{F}}_{I}$ (Sec. \ref{GFPMSS}); 4) Predict the probabilistic mask $\hat{\phi}$ and 
    affordance category $\hat{y}$  (Sec. \ref{final_decoder}).}
    \label{framework}
    \vspace{-1em}
\end{figure*}

Cross-modal learning between image and point cloud data has been widely applied in various computer vision tasks, including point cloud registration \cite{lin2017novel, ren2022corri2p, li2021deepi2p} and object recognition \cite{Qi_2020_CVPR, Lu_2023_CVPR, chen2023pimae}. Images provide rich semantic information such as object categories, textures, and colors, while point clouds capture precise 3D geometric structures of objects. Fusing these modalities enhances scene reconstruction \cite{huang2023cross} and improves object localization \cite{wang2023net, cai2023objectfusion}.

Despite these advancements, leveraging semantic information from images to enhance the understanding of interactions in point clouds remains a critical challenge. Several attempts have been made to address this issue \cite{10374120, Jian_2023_ICCV, 10321669}. Existing approaches typically rely on datasets like 3D AffordanceNet \cite{deng20213d} to pair images with point clouds for identifying interactive regions of objects \cite{gao2024learning2dinvariantaffordance}. For instance, Luo et al. \cite{luo2022learning} capture affordance knowledge from exocentric human-object interactions and utilize a cross-view knowledge transfer framework for grounding. Additionally, LEMON \cite{yang2024lemon} exploits the correlation between interaction counterparts to jointly anticipate human contact, object affordance, and human-object spatial relations in 3D space. Although these studies have advanced geometric understanding and the detection of interactive regions, they primarily focus on identifying and localizing these areas, often neglecting a deeper functional understanding of their roles.

\section{Method}
\subsection{Overview}
Our task is to ground the affordance depicted in an image onto a point cloud representation of the same object category and predict the affordance class. Formally, the input to our method consists of a 3D point cloud $P \in \mathbb{R}^{N \times 3}$ of the object, where $N$ is the number of points, an RGB image $I \in \mathbb{R}^{3 \times H \times W}$ with height $H$ and width $W$ which reflects an affordance category and a pair of bounding boxes $\mathcal{B} = \{B_{\text{sub}}, B_{\text{obj}}\}$ for the affordance-related subject and object entities in the image $I$, respectively. The scene context outside these bounding boxes is denoted by a scene mask $M_{\text{sce}}$. 
Our goal is to predict the affordance category $y$ and generate a point-wise probabilistic mask $\phi \in [0,1]^{N}$, where the probability value at each point represents the likelihood of that point being associated with the affordance. 

To address this task, we propose a multiscale cross-modal fusion framework that aims to generate better coverage for the affordance grounding and consistent prediction on the affordance category. To achieve this, our framework learns an effective affordance-aware 3D representation and a stage-wise inference strategy that exploits the dependency of the grounding and classification subtasks. Specifically, we design an end-to-end network architecture consisting of four main modules, as summarized in Fig. \ref{framework}: 1) \textit{Multi-modal Feature Module}, which extracts context-aware affordance cue from the image and multi-scale geometric features from the point cloud (Sec.~\ref{feature_extra}); 2) \textit{Cross-modal Fusion Module}, which fuses those two modalities to obtain multi-scale affordance-aware 3D representations at the global and local region levels (Sec.~\ref{CMFM}); 3) \textit{Propagation and Selection Module}, which then propagates the local features at each scale via a graph neural network and selects a relevant scale to generate final global and local representations (Sec.~\ref{GFPMSS}); 4) \textit{Affordance Prediction Module}, which first predicts the grounding probabilistic mask and then fuses the global contextual feature and masked local feature to predict the affordance class (Sec.~\ref{final_decoder}). Below, we will introduce the design of each module and the model training (Sec.~\ref{loss_function}) in detail. \textcolor{black}{Additionally, we provide further details on model implementation and training in Appendix Sec.~\ref{method details} and Sec.~\ref{training details}.}

\subsection{Multi-modal Feature Module}\label{feature_extra}
\subsubsection{2D Context-Aware Affordance Feature}\label{2d_extractor}
Given the input image $I$ and the bounding boxes $\mathcal{B} = \{B_{\text{sub}}, B_{\text{obj}}\}$, we first extract an affordance feature encoding the functionality of the target object (indicated by $B_{\text{obj}}$) and its image context. To this end, we first use a vision backbone network to compute an image embedding, followed by ROI-align \cite{he2017mask} to generate object entity feature map $\mathbf{F}_{ob}$,  subject entity feature map $\mathbf{F}_{sb}$, and scene context feature map $\mathbf{F}_{sc}$. We then introduce an adaptive fusion method to integrate those features into a context-aware affordance representation. Specifically, we adopt a cross-attention operator to fuse the object entity with the subject entity and the scene context as follows~\cite{10378483},
\begin{equation}
\begin{gathered}
\mathbf{F}_{e}=\operatorname{MHA}(\mathbf{W}_q\mathbf{F}_{ob}, \mathbf{W}_k\mathbf{F}_{sb}, \mathbf{W}_v\mathbf{F}_{sb}),\\
\mathbf{F}_{s}=\operatorname{MHA}(\mathbf{W}_q\mathbf{F}_{ob}, \mathbf{W}_k\mathbf{F}_{sc}, \mathbf{W}_v\mathbf{F}_{sc}),
\end{gathered}
\end{equation}
where $\mathbf{F}_{e}$ and $\mathbf{F}_{s}$ indicate fused entity and scene features, respectively, $\operatorname{MHA}(Q,K,V)$ represents the standard attention operator with input query $Q$, key $K$ and value $V$, and $\mathbf{W}_q$, $\mathbf{W}_k$ and $\mathbf{W}_v$ are weight matrices. 


We observe that the scene context plays a varying role in affordance prediction and may introduce distraction in certain cases. Therefore, we design a gate operator to selectively integrate scene context into the fused entity features, which generates the final affordance feature $\mathbf{F}_{I}$. Formally, 
\begin{equation}
\mathbf{F}_{I}=\operatorname{LP}\left(\mathbf{F}_{e}+\operatorname{Sigmoid}\left(\operatorname{LN}\left(\mathbf{F}_{s}\right)\right) \odot \mathbf{F}_{s}\right),
\end{equation}
where $\operatorname{Sigmoid}\left(\cdot\right)$ denotes the sigmoid activation function, $\operatorname{LN}\left(\cdot\right)$ is a linear layer, and $\odot$ represents element-wise multiplication. Here $\operatorname{LP}\left(\cdot\right)$ is a linear layer with ReLU activation for the feature fusion.



\subsubsection{Multi-Scale 3D Geometric Feature}
Leveraging multi-scale features is essential for affordance learning. Consider the example of a bag, which exhibits distinct affordances such as lifting and containing. The spatial extent of the affordance associated with lifting is notably smaller compared to that of containing. 

In light of this, we utilize a point cloud backbone network to extract region features at two different scales. Specifically, we first form two types of regions by a 3D backbone network and downsampling operations, and then perform feature pooling in each region, which outputs $\mathbf{F}_P^l \in \mathbb{R}^{C \times N_p^l}$ and $\mathbf{F}_P^s \in \mathbb{R}^{C \times N_p^s}$, where $N_p^l$ and $N_p^s$ are the number of regions at a large and a small scale, respectively. 

Moreover, we construct a similarity graph among regions of each scale based on the point cloud features, and the resulting graph weight matrices, denoted as $\mathbf{S}^l$ and $\mathbf{S}^s$, are computed as follows: 
\begin{equation}
\begin{gathered}
\mathbf{S}^l_{i,j} = \langle(\mathbf{F}_P^l(i), \mathbf{F}_P^l(j)\rangle,\quad
\mathbf{S}^s_{i,j} = \langle(\mathbf{F}_P^s(i), \mathbf{F}_P^s(j)\rangle,
\end{gathered}
\end{equation}
where $\langle\cdot, \cdot\rangle$ denotes the inner product, and $\mathbf{F}_P^{l/s}(i)$ indicate the $i$-th column of $\mathbf{F}_P^{l/s}$.

\subsection{Cross-modal Fusion Module}\label{CMFM}
One of the key challenges in affordance grounding is to effectively align the image-based affordance cue to the point cloud representation of target objects. 
To tackle this, we adopt a simple yet efficient fusion mechanism to integrate features from two modalities. Specifically, we use the multi-head cross-attention to capture various aspects of semantic and geometric information contained in the image and point cloud \cite{park2020mhsan}. Given the context-aware affordance feature $\mathbf{F}_{I}$, we first enhance the multi-scale geometric features of point cloud $\mathbf{F}_P^l$ and $\mathbf{F}_P^s$ using the affordance cues as follows:
\begin{equation}
\begin{gathered}
\mathbf{\tilde{F}}_P^l=\operatorname{MHA}(\mathbf{W}^{l}_q\mathbf{F}_P^{l}, \mathbf{W}^l_k \mathbf{F}_{I}, \mathbf{W}^l_v \mathbf{F}_{I}),\\
\mathbf{\tilde{F}}_P^{s}=\operatorname{MHA}(\mathbf{W}^{s}_q\mathbf{F}_P^{s}, \mathbf{W}^s_k \mathbf{F}_{I}, \mathbf{W}^s_v \mathbf{F}_{I}),
\end{gathered}
\end{equation}
where $\operatorname{MHA}$ is the multi-head attention operator, and $\mathbf{W}^{l/s}_{q,k,v}$ are the weight matrices for its query, key and value.


After obtaining the local region representations $\mathbf{\tilde{F}}_P^l$ and $\mathbf{\tilde{F}}_P^s$, 
we further compute a global affordance representation at each scale by fusing the 3D features into $\mathbf{F}_{I}$ through a similar cross-attention process, which are denoted as $\tilde{\mathbf{F}}_I^l$ and $\tilde{\mathbf{F}}_I^s$.

\subsection{Propagation and Selection Module}\label{GFPMSS}
The affordance regions within an image often represent only a fraction of the object’s full potential affordance area. For instance, the “open” affordance associated with a microwave handle typically exhibits only limited interaction points within the image context, as users generally engage with a specific portion of it. To extrapolate the complete affordance regions within the point cloud, we leverage geometric similarities across various regions of the object to encourage similar regions sharing similar affordance representations, thereby facilitating more accurate grounding. 

Formally, we introduce two-layer GCN networks to refine the multiscale regional features $\mathbf{\tilde{F}}_P^l$ and $\mathbf{\tilde{F}}_P^s$. Specifically, at each scale, we take the similarity graph $\mathbf{S}^l$ (or $\mathbf{S}^s$) and the corresponding features $\mathbf{\tilde{F}}_P^l$ (or $\mathbf{\tilde{F}}_P^s$) as input and iteratively update the features based on message propagation: 
\begin{equation}
\mathbf{R}^{(t+1)} = \operatorname{Sigmoid}\left(\mathbf{\tilde{A}} \mathbf{R}^{(t)} \mathbf{W}^{(t)}\right), \quad t=0,1,2, 
\end{equation}
where $\mathbf{\tilde{A}} = \mathbf{D}^{-\frac{1}{2}} \mathbf{A} \mathbf{D}^{-\frac{1}{2}}$ is the normalized adjacency matrix and  $\mathbf{A}$ is $\mathbf{S}_P^l$ or $\mathbf{S}_P^s$, $ \mathbf{R}^{(t)}$ is the updated regional feature and $\mathbf{W}^{(t)}$ is the learnable weight matrix at layer $t$. The initial regional feature $ \mathbf{R}^{(0)}=\tilde{\mathbf{F}}_P^{l/s} \odot \Gamma (\mathbf{F}_P^{l/s})$, where $\Gamma(\cdot)$ is the pooling and expansion process for feature reweighting.  
We take the GCN outputs, denoted as $\tilde{\mathbf{R}}_P^{l/s}$, as the refined regional affordance features. Subsequently, we select a suitable scale in a soft-weighting manner via a gate fusion network as below
\begin{equation}
  \tilde{\mathbf{R}}_{P}=\alpha\mathbf{U}\left(\tilde{\mathbf{R}}_P^l\right)+(1-\alpha)\mathbf{U}\left(\tilde{\mathbf{R}}_P^s\right),
\end{equation}
where $\mathbf{U}(\cdot)$ is the up-sample operation, $\alpha=\frac{\alpha_1}{\alpha_1+\alpha_2}$, $\alpha_1$ and $\alpha_2$ is a gate network score using $\mathbf{\tilde{R}}_P^l$ and $\mathbf{\tilde{R}}_P^s$ as inputs.
Similarly, we use the same scores to compute the global affordance features as
\begin{equation}
  \tilde{\mathbf{F}}_{I}=\alpha \tilde{\mathbf{F}}_I^l+(1-\alpha) \tilde{\mathbf{F}}_I^s.
\end{equation}

\subsection{Affordance Prediction Module}\label{final_decoder}

\begin{table*}[htbp]
\centering
 \caption{The experimental results of our proposed method and the competitors on PIAD. The best results are marked in bold. Seen and Unseen
are two partitions of the dataset. AUC, aIOU, and ACC are shown in percentage.}
	\label{main_table}
\begin{tabular}{c|c c c c :c|c c c c :c}
\toprule
\multirow{2}{*}{\textbf{Method}} & \multicolumn{5}{c|}{\textbf{Seen}} & \multicolumn{5}{c}{\textbf{Unseen}} \\
 & \textbf{AUC} $\uparrow$ & \textbf{aIOU} $\uparrow$ & \textbf{SIM} $\uparrow$ & \textbf{MAE} $\downarrow$& \textbf{ACC} $\uparrow$& \textbf{AUC} $\uparrow$& \textbf{aIOU} $\uparrow$& \textbf{SIM} $\uparrow$& \textbf{MAE} $\downarrow$& \textbf{ACC} $\uparrow$\\
\hline
Baseline& 82.92& 17.45& 0.522& 0.101& 84.19& 66.09& 6.76& 0.312& 0.138& 29.65\\
XMF& 80.39& 14.42& 0.497& 0.114& 75.40& 67.68& 4.90& 0.303& 0.128& 20.14\\
Laso& 84.26& 19.83& 0.547& 0.093& 73.42& 70.62& 6.00& 0.346& 0.126& 38.33\\
 IAG& 85.96& 20.13& 0.564& 0.092& 83.50& 72.20& 7.39& 0.349& 0.125&42.36\\
\hline
 Ours& \textbf{87.20}& \textbf{22.75}& \textbf{0.604}& \textbf{0.081}& \textbf{90.91}& \textbf{74.40}& \textbf{8.50}& \textbf{0.363}& \textbf{0.117}&\textbf{45.14}\\
\bottomrule
\end{tabular}
\end{table*}

Finally, we use the multiscale affordance features to decode a probabilistic mask via a mask head and its category via a classifier head. As the affordance area is strongly correlated to its category, we adopt a two-stage strategy to generate the final prediction. Specifically, we first predict the grounding based on the local regional features as follows,
\begin{equation}
\hat{\phi}=f_\phi\left(\tilde{\mathbf{R}}_{P}\right).  
\end{equation}
where $f_\phi$ is an MLP taking each local region's feature as input and predicts affordance confidence for that region. 

Given the grounding output, we then combine the global context-aware affordance feature $\hat{{\mathbf{F}}}_{I}$ with the local affordance features to predict the final affordance class, which is formulated as:
\begin{equation}
\hat{y} = f_y\left([P(\tilde{\mathbf{F}}_{I}),P(\hat{\phi} \cdot \tilde{\mathbf{R}}_{P})]\right),
\end{equation}
where $P(\cdot)$ denotes the average pooling operation and $f_y$ is an MLP classifier.
\subsection{Loss Function}\label{loss_function}
While our method employs a sequential prediction strategy, all four main modules in the proposed network are differentiable and hence allow us to train the full model in an end-to-end manner. Our loss function is a combination of grounding loss and classification loss:
\begin{equation}
\mathcal{L} = \mathcal{L}_{g} + \lambda_c \mathcal{L}_{c},
\end{equation}
where \(\mathcal{L}_{g}\) is the grounding loss that measures the discrepancy between the ground truth \(\phi\) and the predicted grounding result \(\hat{\phi}\), computed using a focal loss \cite{ross2017focal} combined with a Dice loss \cite{milletari2016v}. The term \(\mathcal{L}_{c}\) denotes the cross-entropy loss for classification and \(\lambda_c\) is a hyperparameter that balances the two terms.

\section{Experiment}
\subsection{Experimental Setting}
\subsubsection{Dataset and baselines}
\begin{table*}[h]
\centering
\caption{The ablation study to investigate the effectiveness of main modules of our method. The best results are marked in bold.}
	\label{ablation_study}
\begin{tabular}{c|cccc:c|cccc:c}
\toprule
\multirow{2}{*}{\textbf{Variants}} & \multicolumn{5}{c|}{\textbf{Seen}}          & \multicolumn{5}{c}{\textbf{Unseen}}        \\ 
                  & \textbf{AUC} $\uparrow$ & \textbf{aIOU} $\uparrow$ & \textbf{SIM} $\uparrow$ & \textbf{MAE} $\downarrow$& \textbf{ACC} $\uparrow$& \textbf{AUC} $\uparrow$& \textbf{aIOU} $\uparrow$& \textbf{SIM} $\uparrow$& \textbf{MAE} $\downarrow$& \textbf{ACC} $\uparrow$\\ \hline
Blind         &              82.92&               17.45&             0.522&              0.101&              84.19&              66.09&               6.76&             0.312&              0.138&              29.65\\

 w/o MSI& 85.96& 20.39& 0.579& 0.087& 88.93& 70.98& 7.13& 0.359& 0.130&34.40\\
 w/o GFPM& 85.83& 19.93& 0.581& 0.092& 90.32& 70.69& 6.79& 0.348& 0.139&38.43\\
 w/o CGC& 86.49& 22.06& 0.593& 0.084& 90.02& 70.61& 6.25& 0.337& 0.148&38.33\\  
w/o SG             & 86.30&               21.58&             0.599&              0.090&              89.92&              71.42&               6.54&             0.355&              0.119&              41.74\\  
\hline
Ours&              \textbf{87.20}&               \textbf{22.75}&             \textbf{0.604}&              \textbf{0.081}&              \textbf{90.91}&              \textbf{74.40}&               \textbf{8.50}&             \textbf{0.363}&              \textbf{0.117}&              \textbf{45.14}\\  
\bottomrule
\end{tabular}

\end{table*}

Given that few studies simultaneously address both affordance grounding and classification, public datasets on this topic are limited. In our paper, we utilize the PIAD dataset collected in \cite{10378483} for comparison. This dataset contains 7,012 point cloud instances and 5,162 images across 23 object categories and 17 affordance classes. Each point cloud in the image-point cloud pairs consists of 2,048 points. \textcolor{black}{Notably, compared to existing affordance datasets, PIAD offers a similar breadth of affordance categories, object classes, and instances while standing out as the dataset with the largest number of images, as detailed in Table~\ref{dataset_comparison}.} We adopt two experimental settings: seen and unseen. In the seen setting, the training and test sets share similar class categories, while in the unseen setting, some object classes in the test set may not appear in the training set.

We compare our method with XMF \cite{aiello2022cross}, Laso \cite{li2024laso}, and IAG \cite{10378483}. As a baseline, we extract features directly and fuse them using multi-head attention, then concatenate the multi-modal features to serve as input to two MLP networks for grounding and classification results. To ensure fairness, all compared methods use the same image backbone, Swin \cite{liu2021swin}, and the point cloud backbone, PointNet++ \cite{qi2017pointnet++}. Details of these baselines are provided in the Appendix Sec~\ref{baselines}. Additionally, considering that IAG originally utilizes ResNet to extract image features, we also compare our method with it via ResNet in the Appendix Sec~\ref{resnet}.

\begin{table}[h]
    \caption{Comparison of different affordance datasets in terms of object class, affordance class, point cloud count, text annotations, and image samples.}
    \label{dataset_comparison}
    \centering
    \resizebox{\columnwidth}{!}{ 
        \begin{tabular}{lccccc}
            \toprule
            \textbf{Method} & \textbf{Object CLS} & \textbf{Affordance CLS} & \textbf{Instance} & \textbf{Text} & \textbf{Image} \\
            \midrule
            AdaAfford \cite{wang2022adaafford}& 15 & 2 & 972 & - & - \\
            Where2Explore \cite{NEURIPS2023_0e7e2af2} & 14 & 2 & 942 & - & - \\
            LASO \cite{li2024laso}& 23 & 18 & 8434 & 870 & - \\
            Lemon \cite{yang2024lemon}& 21 & 17 & 5143 & - & 5001 \\
            Ours & 23 & 17 & 7012 & - & 5162 \\
            \bottomrule
        \end{tabular}
    }
\end{table}

\subsubsection{Evaluation metrics}
Most methods for locating affordance areas utilize four metrics for comparison \cite{deng20213d,nagarajan2019grounded,zhai2022one}, specifically AUC, aIOU, SIM, and MAE. For affordance classification, we report the accuracy (ACC) results. Except for MAE, higher values for all other metrics indicate better performance. \textcolor{black}{Additionally, detailed calculations of the evaluation metrics are provided in Appendix Sec.~\ref{evaluation metrics}.}

\subsection{Experimental results}
The experimental results of our method and those of the competitors are reported in Table~\ref{main_table}. The table demonstrates that our method outperforms all compared methods across all metrics. For instance, it surpasses the second-best algorithm by 1.44\%, 13.02\%, 7.09\%, 11.96\%, and 7.98\% on the five metrics in the Seen setting, respectively. Significant improvements are also observed in the Unseen setting, which verifies the effectiveness of our proposed method.

Additionally, both IAG and our method integrate affordance classification and grounding simultaneously, which may explain their superior performance and inspire future research. However, IAG performs worse in affordance classification. We attribute this to its failure to effectively couple affordance classification with affordance grounding. Furthermore, while Laso originally utilized text information to guide grounding, we replace its text backbone with an image backbone to address the image-guided 3D affordance grounding problem, yielding acceptable results. Therefore, in the future, we could propose a method capable of handling both text-guided and image-guided 3D affordance tasks. \textcolor{black}{Furthermore, we compare model sizes to demonstrate the scalability of our approach. The results are provided in Appendix Sec.\ref{modelsize}. Similarly, further details on hyperparameter analysis experiments are provided in Appendix Sec.~\ref{hyper}.}



\begin{figure*}
    \centering
    \includegraphics[width=1\linewidth]{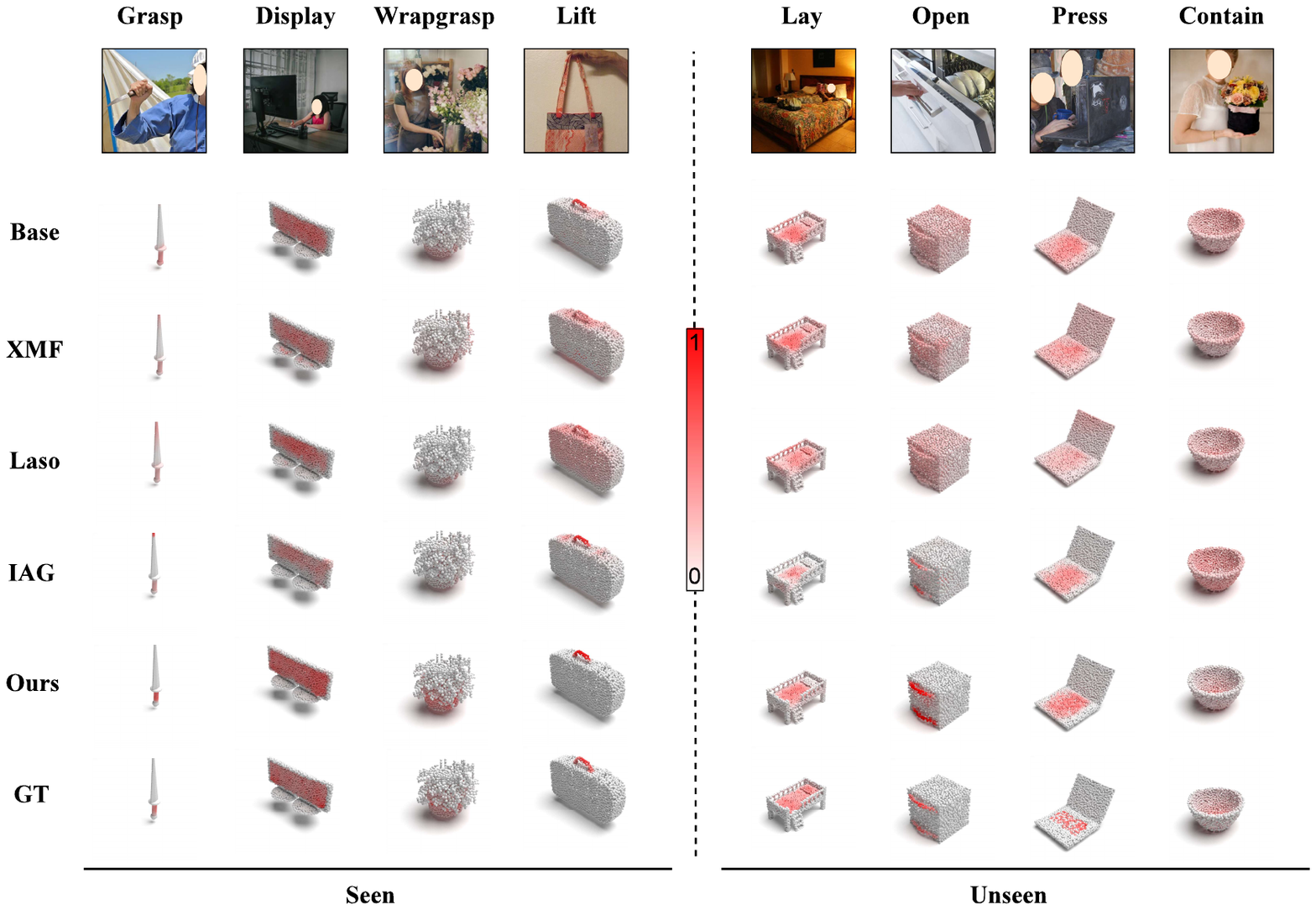}
    \caption{The visualization results. The first row displays the interactive images that reflect object affordances. The last row shows the ground truth of the affordance regions in the 3D point cloud. The left four columns correspond to the Seen setting, while the right four columns represent the Unseen setting.}
    \label{via_compare}
    \vspace{-1em}
\end{figure*}
\subsection{Ablation study}
To verify the effectiveness of different modules in our model, we conduct an ablation study by removing one module at a time. The main contributions of our model can be summarized in four parts: multi-scale information (MSI), geometric feature propagation module (GFPM), coupling grounding and classification (CGC), and scene gate (SG).

For MSI, we extract region features at a fixed scale and remove the multi-scale fusion module to assess its effectiveness. We perform similar operations to evaluate the impact of GFPM and SG. For CGC, we couple 3D affordance grounding and affordance classification using a grounding-classification pipeline, and we examine the effects of removing this module by simultaneously outputting the results. The term 'Blind' refers to the results obtained by removing all key contributions. 

The results are documented in Table~\ref{ablation_study}. From the table, we observe that performance declines when any of the main contributions are removed from our model, particularly for MSI and GFPM. 

\textcolor{black}{Additionally, to further illustrate the effectiveness of these components, we present visualizations of ablation study in Section~\ref{sec:vis_ablation} and Fig.~\ref{via_ablation} below. These visual results provide an intuitive understanding of how each module impacts our tasks.}

\subsection{Visualization}
\textcolor{black}{In previous sections, we demonstrated the superiority of our model through statistical analysis. In this section, we provide visualization results to further illustrate the effectiveness of our method. Specifically, we compare our model with competing approaches and present visualization of ablation study to validate key components. Additionally, we analyze the effect of varying object instances for a given image and different affordance depictions for a given object, with detailed results provided in Appendix Sec~\ref{different instances} and Sec~\ref{different affordances} for completeness.}
\subsubsection{Comparison results}

We present the visualization results of our method and the competing approaches in Fig. ~\ref{via_compare}. The figure clearly illustrates that our proposed method delivers superior performance in terms of visualization quality. Moreover, the experimental results highlight the importance of coupling affordance grounding with affordance classification, region propagation, and multi-scale information.
1)In the Seen setting, a knife has three affordances: cut, grasp, and stab. When reasoning with images depicting grasping, our competitors mistakenly identify both the grasping and stabbing regions simultaneously, leading to confusion. This highlights their failure to effectively couple affordance areas with their corresponding functionalities. In contrast, our approach successfully couples these elements, addressing this issue.
2)For the lay affordance of a bed, existing methods tend to predict incomplete areas. Our geometric feature propagation technique effectively addresses this issue, enabling the prediction of the complete affordance area.
3)Regarding the lifting affordance of a bag, which corresponds to regions occupying a relatively small part of the object, using larger-scale segmentation unavoidably affects surrounding areas. Our multi-scale selection method effectively overcomes this challenge, resulting in a more accurate prediction of the affordance area.
\textcolor{black}{Additional extensive comparison results are provided in Appendix Sec.~\ref{more comparison results}.}


\subsubsection{Visualization of Ablation Study}
\label{sec:vis_ablation}
\begin{figure}
    \centering
    \includegraphics[width=1\linewidth]{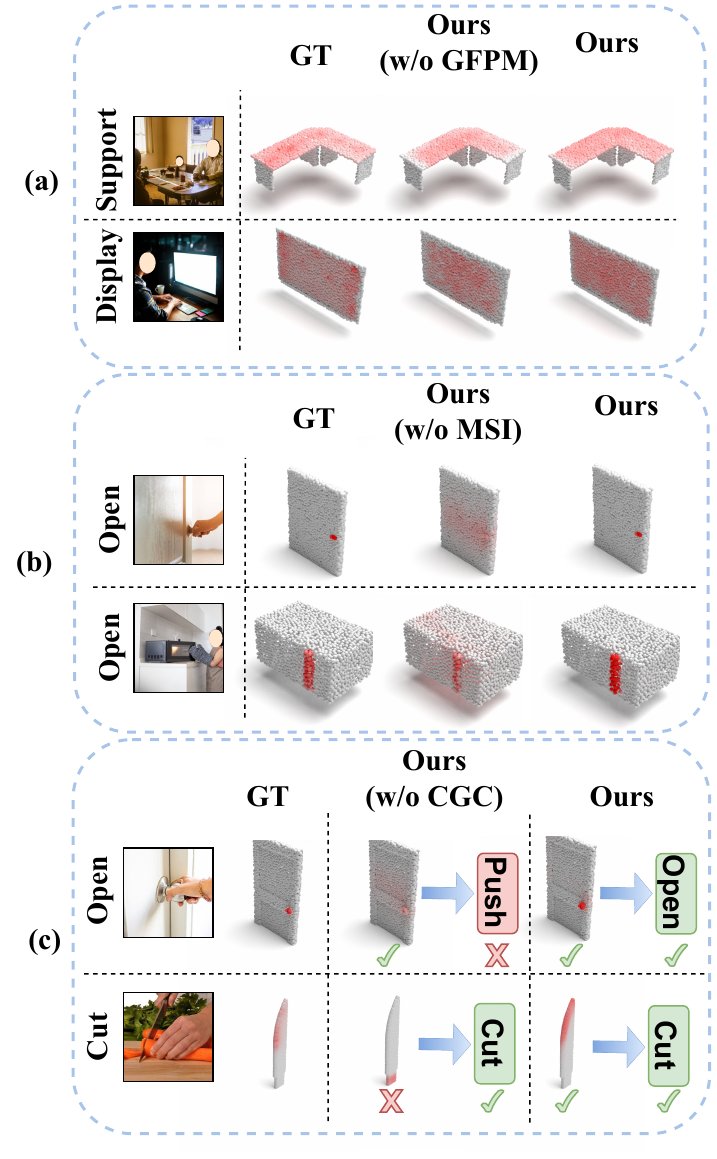}
    \caption{Visualization of ablation studies: (a) Without GFPM, predictions fail to cover the full affordance area. (b) Removing MSI reduces precision for fine-grained regions. (c) Without CGC, grounding areas are confused with incorrect functionalities.}
    \label{via_ablation}
    \vspace{-1em}
\end{figure}

\textcolor{black}{To further illustrate the effectiveness of GFPM, MSI and CGC in our method, we visualize ablation results in Fig.~\ref{via_ablation}, showing that: 
1) GFPM (Fig.~\ref{via_ablation}(a)): Without GFPM, grounding predictions fail to cover the full potential affordance area, indicating the importance of geometric feature propagation;
2) MSI (Fig.~\ref{via_ablation}(b)): Removing multi-scale information impairs the model's precision in grounding fine-grained interaction regions;
3) CGC (Fig.~\ref{via_ablation}(c)): Without CGC, the model confuses affordance areas and their functionalities, highlighting the necessity of coupling grounding with classification.}

These visualizations validate the significant contribution of each component to accurate image-guided 3D affordance grounding and classification.

\subsection{Cross-Dataset Generalization }
\begin{figure}
    \centering
    \includegraphics[width=0.9\linewidth]{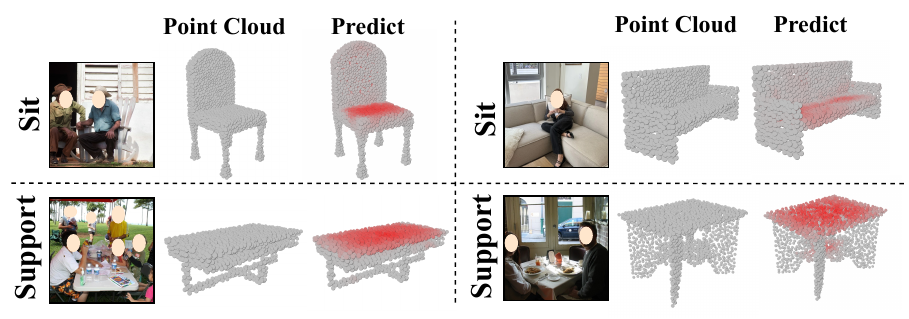}
    \caption{Visualization of cross-dataset generalization results on ShapeNet point clouds }
    \label{cross_dataset}
    \vspace{-1em}
\end{figure}

\textcolor{black}{To further validate the generalization capability of our model, we collected additional point clouds from other datasets and evaluated our approach on these samples. Fig.~\ref{cross_dataset} visualizes our grounding results on point cloud samples from the ShapeNet dataset \cite{chang2015shapenetinformationrich3dmodel}. As illustrated, our model achieves favorable grounding results, demonstrating strong generalization across datasets. Notably, the sofa category does not appear in PIAD, yet our model achieves strong grounding performance on such objects, further demonstrating its robustness and generalization across datasets.}

\subsection{Limitations}
\begin{figure}
    \centering
    \includegraphics[width=0.8\linewidth]{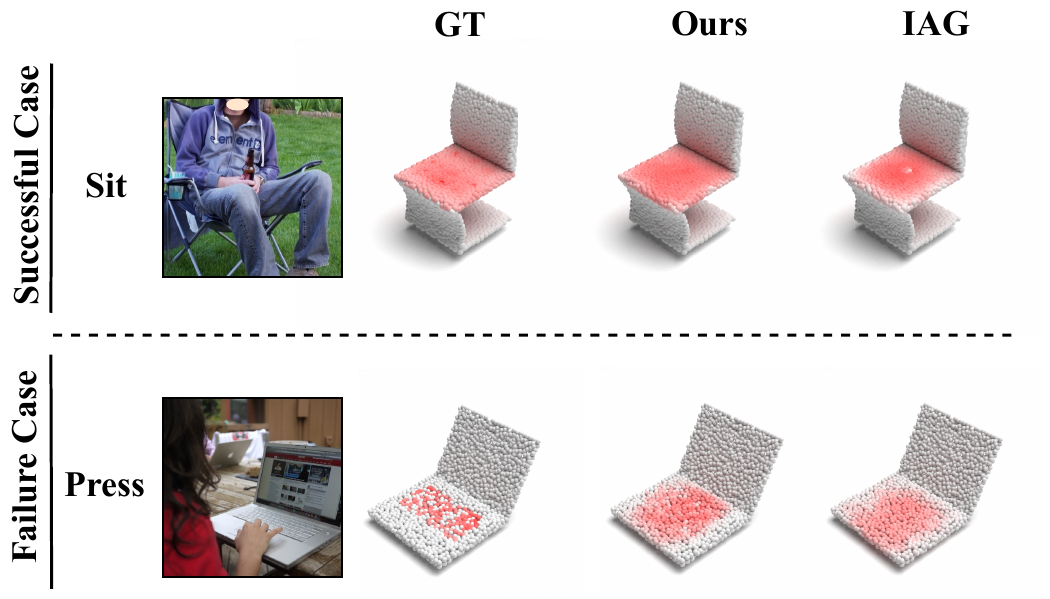}
    \caption{The success and failure cases of our region propagation.}
    \label{case_study}
    \vspace{-1em}
\end{figure}

In our method, we leverage the inherent geometric similarity within point clouds to promote consistent grounding results for geometrically similar regions. However, this approach may occasionally result in predicted areas that extend beyond the true potential affordance areas. As shown in Fig. ~\ref{case_study}, for instance, in the case of the 'press' affordance of a computer keyboard, the predicted area may erroneously extend to regions outside the keyboard. While the geometric feature propagation technique has generally been effective in most experiments, this issue could be due to the poor quality of the point cloud data for the keyboard, making it difficult to accurately distinguish the keyboard from its surrounding areas. To address this, we can collect higher-quality point cloud data for further verification.


\section{Conclusion}
In this work, we propose a novel approach that integrates context-aware affordance information from images to enhance functional region learning in 3D point clouds of the same object category. By combining localized functional regions with global context from 2D images, our method improves 3D affordance understanding. To address fixed-scale learning limitations and incomplete affordance area predictions, we introduce multi-scale feature selection and geometric feature propagation. Our approach not only enhances performance but also paves the way for future advancements in affordance grounding across diverse contexts.
\section{Acknowledgments}
This work was supported by Shanghai Frontiers Science Center of Human-centered Artificial Intelligence, and MoE Key Lab of Intelligent Perception and Human-Machine Collaboration (ShanghaiTech University)



{
    \small
    \bibliographystyle{ieeenat_fullname}
    \bibliography{main}
}

\clearpage
\appendix
\setcounter{section}{0} 
\renewcommand{\thesection}{\Alph{section}} 

\clearpage
\setcounter{page}{1}
\maketitlesupplementary

\section{Implementation Details}

\subsection{Baselines}
\label{baselines}
We selected two 3D Affordance Grounding works, including IAG\cite{10378483} and Laso\cite{li2024laso}, for comparison. IAG and Laso utilize 2D image and textual information respectively to guide 3D affordance grounding. The key difference is that IAG performs both 3D Affordance Grounding and Affordance Classification, while Laso only performs 3D Affordance Grounding. Additionally, we chose a related work, XMF\cite{aiello2022cross}, for comparison. Furthermore, we designed a baseline model without any innovative modules. All comparison methods share the same backbone network as our method. We retained the cross-modal fusion strategies of these methods but adapted their output heads to perform both 3D Affordance Grounding and Affordance Classification tasks. The specific designs are as follows:

\begin{itemize}
    \item Baseline: In the baseline design, we use the same method as our framework to extract context-aware affordance information from the image. For point cloud feature extraction, we adopt the standard PointNet++ method. Then, the features of the image and point cloud are fused using a cross-attention mechanism, and the fused representation is used as input for two output modules: 3D Affordance Grounding and Affordance Classification.
    \item IAG\cite{10378483}: We replace IAG’s image backbone with the same backbone as our method while keeping the rest of IAG's original design unchanged.
    \item Laso\cite{li2024laso}: We replace Laso’s textual input with image input and adopt the same backbone as our method. Meanwhile, we retain Laso’s original design for cross-modal learning mechanisms and cross-modal fusion. To meet the new task requirements, we add an Affordance Classification output module to Laso’s original output modules.
    \item XMF\cite{aiello2022cross}: XMF is used for fusing images and point clouds to complete the point cloud completion task. We replace XMF’s image backbone with the same backbone as our method while retaining XMF’s cross-modal learning design. To meet the new task requirements, we replace XMF's original output modules with the 3D Affordance Grounding and Affordance Classification output modules.
\end{itemize}
\subsection{Method Details}
\label{method details}
\begin{figure}
    \centering
    \includegraphics[width=1\linewidth]{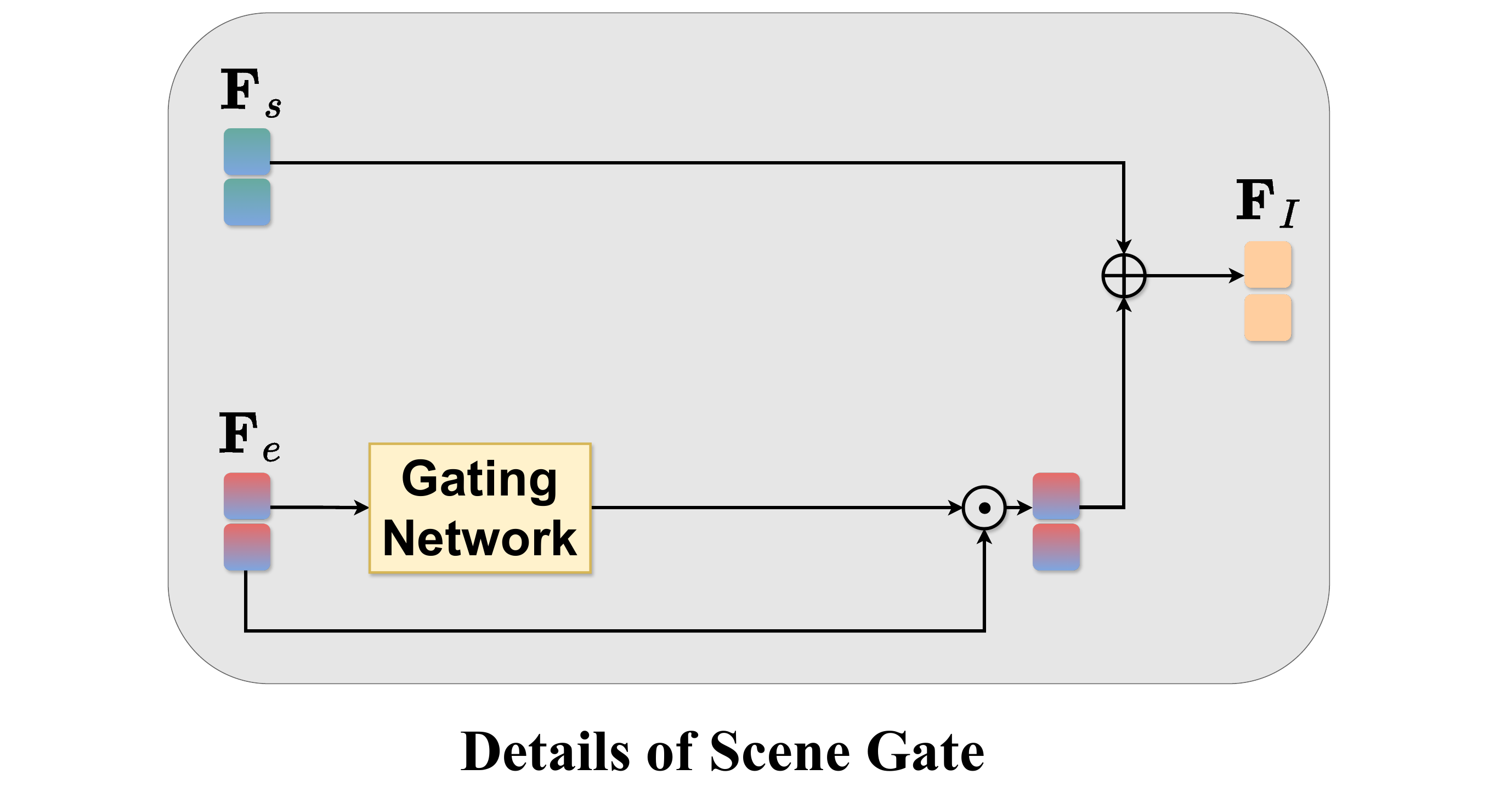}
    \caption{The details of our scene gata.}
    \label{scene_gate}
\end{figure}

\begin{figure}
    \centering
    \includegraphics[width=1\linewidth]{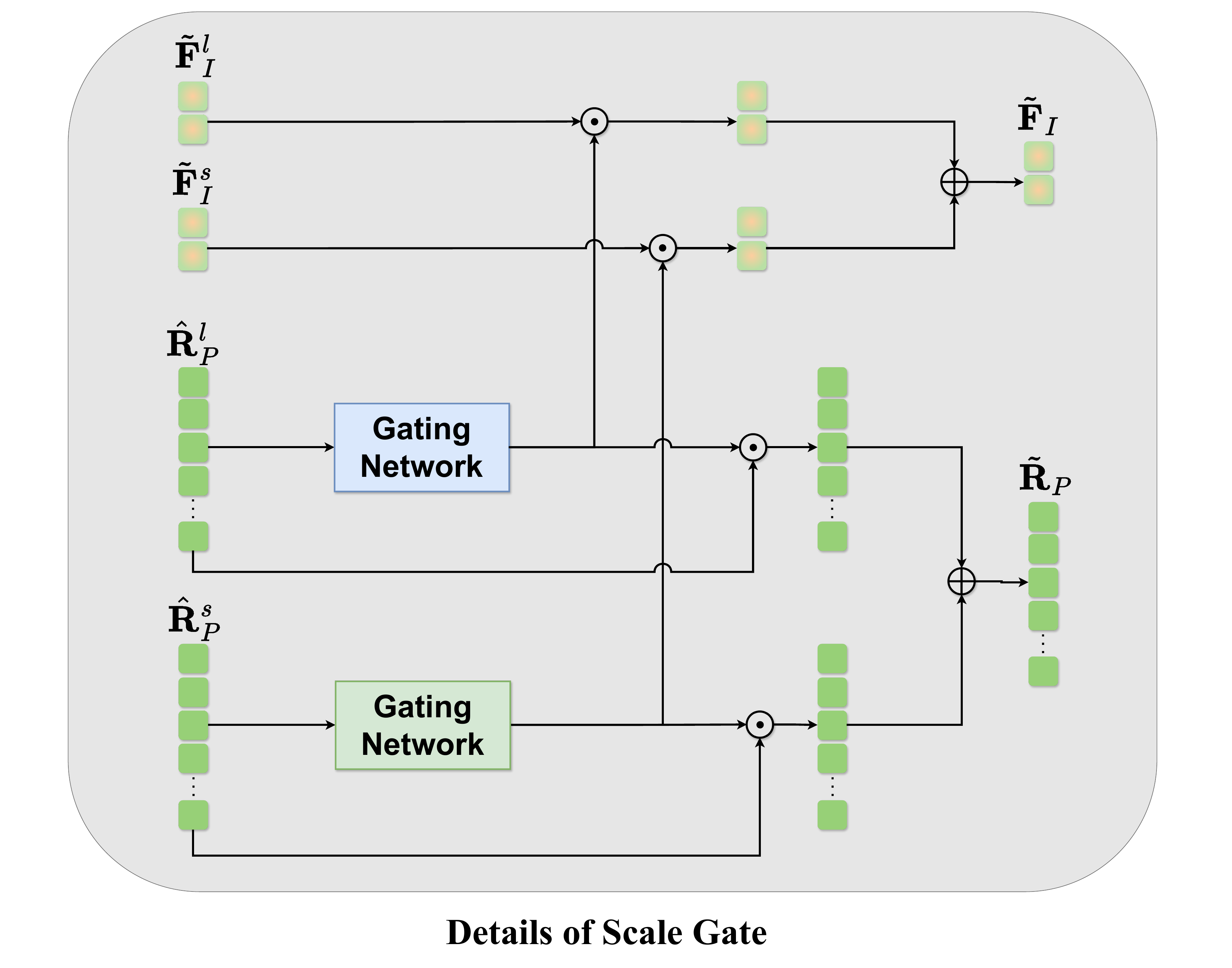}
    \caption{The details of our scale gate.}
    \label{scale_gate}
\end{figure}
We use a pretrained Swin model to extract image features. Since the input images possess varying sizes, inspired by IAG, we first resize the images to $224 \times 224$ and then use the backbone to extract the initial affordance feature $\mathbf{F}_{ini} \in \mathbb{R}^{512 \times 7 \times 7}$. Next, we generate a scene mask $M_{\text{sce}}$ using the bounding boxes of the subject and object entities in the image. The three components are then located in the corresponding positions in $\mathbf{F}_{ini}$, and their features are extracted using ROI alignment to attain object entity feature map $\mathbf{F}_{ob} \in \mathbb{R}^{512 \times 16}$,  subject entity feature map $\mathbf{F}_{sb} \in \mathbb{R}^{512 \times 16}$, and scene context feature map $\mathbf{F}_{sc} \in \mathbb{R}^{512 \times 16}$.

Subsequently, we employ a multi-head attention mechanism (MHA) to fuse the object with both the subject and the scene features to attain $\mathbf{F}_{e} \in \mathbb{R}^{512 \times 16}$ and $\mathbf{F}_{s} \in \mathbb{R}^{512 \times 16}$. A scene gating operation is then applied to selectively integrate the scene information,  as show in Fig.\ref{scene_gate}. After which the two components are fused to obtain 2D context-aware affordance feature $\mathbf{F}_{I} \in \mathbb{R}^{512 \times 16}$. Following this, we use the Pointnet++ and downsampling operations on the point cloud to extract multi-scale features $\mathbf{F}_P^l \in \mathbb{R}^{512 \times 64}$ and $\mathbf{F}_P^s \in \mathbb{R}^{512 \times 128}$, and construct graphs for different scales among regions to attain $\mathbf{S}^l \in \mathbb{R}^{64 \times 64}$ and $\mathbf{S}^s \in \mathbb{R}^{128 \times 128}$.

Next, in the cross-modal fusion module, we also use MHA  to obtain cross-modal representations. Specifically, we first enhance the point cloud features using the context-aware affordance feature, resulting in $\mathbf{\tilde{F}}_P^l \in \mathbb{R}^{512 \times 64}$ and $\mathbf{\tilde{F}}_P^s \in \mathbb{R}^{512 \times 128}$. Then, the updated multi-scale geometric region features are used to refine the multi-scale global contextual features $\tilde{\mathbf{F}}_I^l \in \mathbb{R}^{512 \times 16}$ and $\tilde{\mathbf{F}}_I^s \in \mathbb{R}^{512 \times 16}$.

After the cross-modal fusion, we perform multi-scale decoding operations. We propagate regional information using a two-layer GCN network. The input to the GCN is $\tilde{\mathbf{F}}_P^{l/s} \odot \Gamma (\mathbf{F}_P^{l/s})$ and $\mathbf{S}^{l/s}$, and the learnable weight matrix of each layer has the same dimensions $\mathbb{R}^{512 \times 512}$. After obtaining the outputs $\tilde{\mathbf{R}}_P^{l} \in \mathbb{R}^{512 \times 64}$ and $\tilde{\mathbf{R}}_P^{s} \in \mathbb{R}^{512 \times 128}$, as the refined regional affordance features, we combine these with the earlier downsampling parameters to perform upsampling, then a gating network is then used for soft selection of the multi-scale features as $\tilde{\mathbf{R}}_{P} \in \mathbb{R}^{512 \times 2048}$, which are fed into an MLP network to produce the grounding result $\hat{\phi} \in \mathbb{R}^{2048 \times 1}$. At the same time, we share the selection weights of semantic and geometric features and transfer these weights to the semantic features, obtaining $\tilde{\mathbf{F}}_{I} \in \mathbb{R}^{512 \times 16}$, the details are summarized in Fig. \ref{scale_gate}.

Finally, we use the grounding result as a weight on $\tilde{\mathbf{R}}_{P}$. Then conduct the average pooling on it and $\tilde{\mathbf{F}}_{I}$, we finally concatenate them as the input of our affordance class head to predict the affordance category $\hat{y}$. The notation table is summarized in Table \ref{notaion}.
\begin{table*}[h]
\caption{The dimension and meaning of the tensors of our method.}
\centering
\begin{tabular}{lll}
\toprule
Tensor & Dimension & Meaning \\
\midrule
$\mathbf{F}_{ini}$ & 512 × 7 × 7 & image extractor output \\
$\mathbf{F}_{ob,sb,sc}$ & 512 × 16 & features output by roi-align \\
$\mathbf{F}_{I}$ & 512 × 16 & context-aware affordance feature \\
$\mathbf{F}_P^{l/s}$ & 512 × 64/128 & large/small-scale  regional feature  \\
$\mathbf{S}^{l/s}$ & 64/128 × 64/128 & similarity graph of large/small-scale regional feature  \\
$\mathbf{\tilde{F}}_P^{l/s}$ & 512 × 64/128 & large/small-scale regional feature after cross-modal fusion \\
$\mathbf{\tilde{F}}_I^{l/s}$ & 512 × 16 & context-aware affordance feature with large/small-scale regional information \\
$\tilde{\mathbf{R}}_P^{l/s}$ & 512 × 64/128 &  large/small-scale regional feature after message propagation \\
$\tilde{\mathbf{R}}_{P}$ & 512 × 2048 & regional feature after scale selection  \\
$\tilde{\mathbf{F}}_{I}$ & 512 × 16 & context-aware affordance feature after scale selection\\
$\hat{\phi}$ & 2048 × 1&predicted grounding result\\
$\hat{y}$ & 1 &predicted affordance category result\\
\bottomrule
\end{tabular}
\label{notaion}
\end{table*}

\subsection{Evaluation Metrics}
\label{evaluation metrics}
Below are five used evaluation metrics and their calculation methods:

\begin{enumerate}
    \item \textbf{AUC (Area Under Curve)}: \\
    AUC \cite{auc} is an evaluation metric for point cloud saliency maps. It represents the area under the ROC (Receiver Operating Characteristic) curve. When predicting saliency maps, AUC treats the saliency map as a binary classifier under different thresholds. By measuring the true positive rate (TPR) and false positive rate (FPR) at each threshold, the ROC curve is plotted, and its area is calculated. A larger AUC value indicates higher accuracy of the model in predicting salient regions.
    
    \item \textbf{aIoU (Average Intersection Over Union)}: \\
    aIoU \cite{aiou} is used to evaluate the overlap between the predicted affordance region and the ground truth region in the point cloud. IoU measures the similarity by calculating the ratio of the intersection to the union of the predicted and ground truth regions. The formula is as follows:
    \begin{equation}
        \text{IoU} = \frac{\text{TP}}{\text{TP} + \text{FP} + \text{FN}},
    \end{equation}
    where \(\text{TP}\) (True Positive) represents the number of correctly predicted positive examples, \(\text{FP}\) (False Positive) represents the number of incorrectly predicted positive examples, and \(\text{FN}\) (False Negative) represents the number of missed positive examples. A higher IoU value indicates a greater similarity between the predicted and ground truth regions.
    
    \item \textbf{SIM (Similarity)}: \\
    SIM \cite{sim} measures the similarity between the predicted saliency map and the ground truth saliency map. For a predicted map \(P\) and a ground truth map \(Q_D\), SIM is computed by taking the minimum value between them at each pixel position and summing these minimum values. Both the predicted map and the ground truth map need to be normalized so that their pixel sums equal 1. The formula for SIM is:
    \begin{equation}
        \text{SIM}(P, Q_D) = \sum_i \min(P_i, Q_{D_i}).
    \end{equation}
    A higher SIM value indicates greater similarity between the prediction and the ground truth.
    
    \item \textbf{MAE (Mean Absolute Error)}: \\
    MAE \cite{mae} evaluates the point-wise error between the predicted saliency map and the ground truth saliency map. It is calculated by taking the sum of the absolute values of all prediction errors and dividing it by the total number of points \(n\):
    \begin{equation}
        \text{MAE} = \frac{1}{n} \sum_{i=1}^n |e_i|,
    \end{equation}
    where \(e_i\) is the error for each point. A smaller MAE value indicates a smaller difference between the prediction and the ground truth.

    \item \textbf{ACC (Accuracy)}: \\
    ACC evaluates the proportion of correctly classified points in the entire dataset. It is defined as the ratio of the sum of true positives (\(\text{TP}\)) and true negatives (\(\text{TN}\)) to the total number of points. The formula is:
    \begin{equation}
        \text{ACC} = \frac{\text{TP} + \text{TN}}{\text{TP} + \text{FP} + \text{FN} + \text{TN}}.
    \end{equation}
    A higher ACC value indicates a better overall classification performance.
\end{enumerate}

\subsection{Training Details}
\label{training details}
Our model is implemented in PyTorch and is trained using the Adam optimizer. We set the number of epochs to 150 and the batch size to 16. To ensure fairness, we run comparative algorithms with both 150 epochs and the recommended values from their respective papers, reporting the highest results. All experiments are conducted on a single A100 GPU with an initial learning rate of 0.0005. Our model includes a hyperparameter to balance the weights of two loss components, set to 0.3. We use Swin as the backbone for images, loading pretrained weights, while the point cloud feature extractor uses an untrained PointNet++. Like IAG, each point cloud is paired with two images during training.

\section{Experiments}

\subsection{Backbones}
\label{resnet}
Considering that IAG uses ResNet as the backbone for image feature extraction, in this section, we compare both IAG and our algorithm using the feature extractor specified in the original IAG paper, while keeping other components unchanged. The results are presented in Table \ref{resnet_backbone}. From the table, it can be observed that our model still outperforms IAG in both affordance grounding and  classification tasks, especially in affordance classification. Therefore, our algorithm demonstrates superior performance across different feature extractors, further validating the effectiveness of our model.
\begin{table*}[htbp]
\centering
 \caption{Comparison of Experimental Results between IAG and Our Proposed Method using ResNet.}
	\label{resnet_backbone}
\begin{tabular}{c|c c c c :c|c c c c :c}
\toprule
\multirow{2}{*}{\textbf{Method}} & \multicolumn{5}{c|}{\textbf{Seen}} & \multicolumn{5}{c}{\textbf{Unseen}} \\
 & \textbf{AUC} $\uparrow$ & \textbf{aIOU} $\uparrow$ & \textbf{SIM} $\uparrow$ & \textbf{MAE} $\downarrow$& \textbf{ACC} $\uparrow$& \textbf{AUC} $\uparrow$& \textbf{aIOU} $\uparrow$& \textbf{SIM} $\uparrow$& \textbf{MAE} $\downarrow$& \textbf{ACC} $\uparrow$\\
\hline
IAG& 84.84& 20.23& 0.561& 0.096& 82.31& 71.67& 7.93& 0.352& 0.130& 33.62\\
Ours& 85.76& 21.33& 0.578& 0.086& 86.36& 72.75& 8.89& 0.364& 0.127& 39.77\\
\bottomrule
\end{tabular}
\end{table*}

\subsection{Model Size}
\label{modelsize}
\textcolor{black}{We introduce multi-scale features and a geometric feature propagation module to address the issue of incomplete affordance area prediction and the lack of multi-scale information utilization in existing methods. However, their inclusion may increase model size. To demonstrate the scalability of our approach, we compare model sizes with existing methods, as shown in Table \ref{model_size_table}. Despite incorporating geometric similarity modeling, our model remains smaller than Laso and XMF and is comparable to IAG, highlighting its efficiency and ability to achieve superior performance without excessive computational cost.}

\begin{table}[h!]
    \caption{Comparison of model sizes (in MB).}
    \label{model_size_table}
    \centering
    \small 
    \begin{tabular}{lcccc}
        \toprule
        \textbf{Model} & \textbf{XMF} & \textbf{LASO} & \textbf{IAG} & \textbf{Ours} \\
        \midrule
        \textbf{Size (MB)} & 392.23 & 391.14 & 366.53 & 370.55 \\
        \bottomrule
    \end{tabular}
\end{table}

\subsection{Hyper-Parameter Analysis}
\label{hyper}
In our model, we use a hyperparameter to balance the relative importance of classification loss and grounding loss. To investigate the impact of this hyperparameter on model performance, we performed a sensitivity analysis, as shown in Table \ref{hyper_analysis}. The results indicate that when the hyper-parameter $\lambda_c$ is set to 0.3, our model achieves the best performance on both the seen and unseen settings. Therefore, we choose 0.3 as the optimal value for this hyper-parameter.
\begin{table*}[htbp]
\centering
\caption{The experimental results with different hyperparameters.}
\label{hyper_analysis}
\begin{tabular}{c|c c c c :c|c c c c :c}
\toprule
\multirow{2}{*}{\textbf{$\lambda$}} & \multicolumn{5}{c|}{\textbf{Seen}} & \multicolumn{5}{c}{\textbf{Unseen}} \\
 & \textbf{AUC} $\uparrow$ & \textbf{aIOU} $\uparrow$ & \textbf{SIM} $\uparrow$ & \textbf{MAE} $\downarrow$ & \textbf{ACC} $\uparrow$ & \textbf{AUC} $\uparrow$ & \textbf{aIOU} $\uparrow$ & \textbf{SIM} $\uparrow$ & \textbf{MAE} $\downarrow$ & \textbf{ACC} $\uparrow$ \\
\hline
0.3 & 87.20 & 22.75 & 0.604 & 0.081 & 90.91 & 74.40 & 8.50 & 0.363 & 0.117 & 45.14 \\
0.1 & 86.88 & 22.34 & 0.591 & 0.083 & 90.71 & 73.40 & 7.86 & 0.355 & 0.122 & 41.94 \\
0.5 & 86.84 & 22.47 & 0.597 & 0.083 & 90.71 & 73.39 & 7.99 & 0.356 & 0.125 & 43.70 \\
0.7 & 86.52 & 21.69 & 0.589 & 0.084 & 90.12 & 72.80 & 7.55 & 0.343 & 0.129 & 42.25 \\
\bottomrule
\end{tabular}
\end{table*}
\subsection{More Visualization Results}
\label{more vis}
\subsubsection{More comparison results}
\label{more comparison results}
\begin{figure*}
    \centering
    \includegraphics[width=1\linewidth]{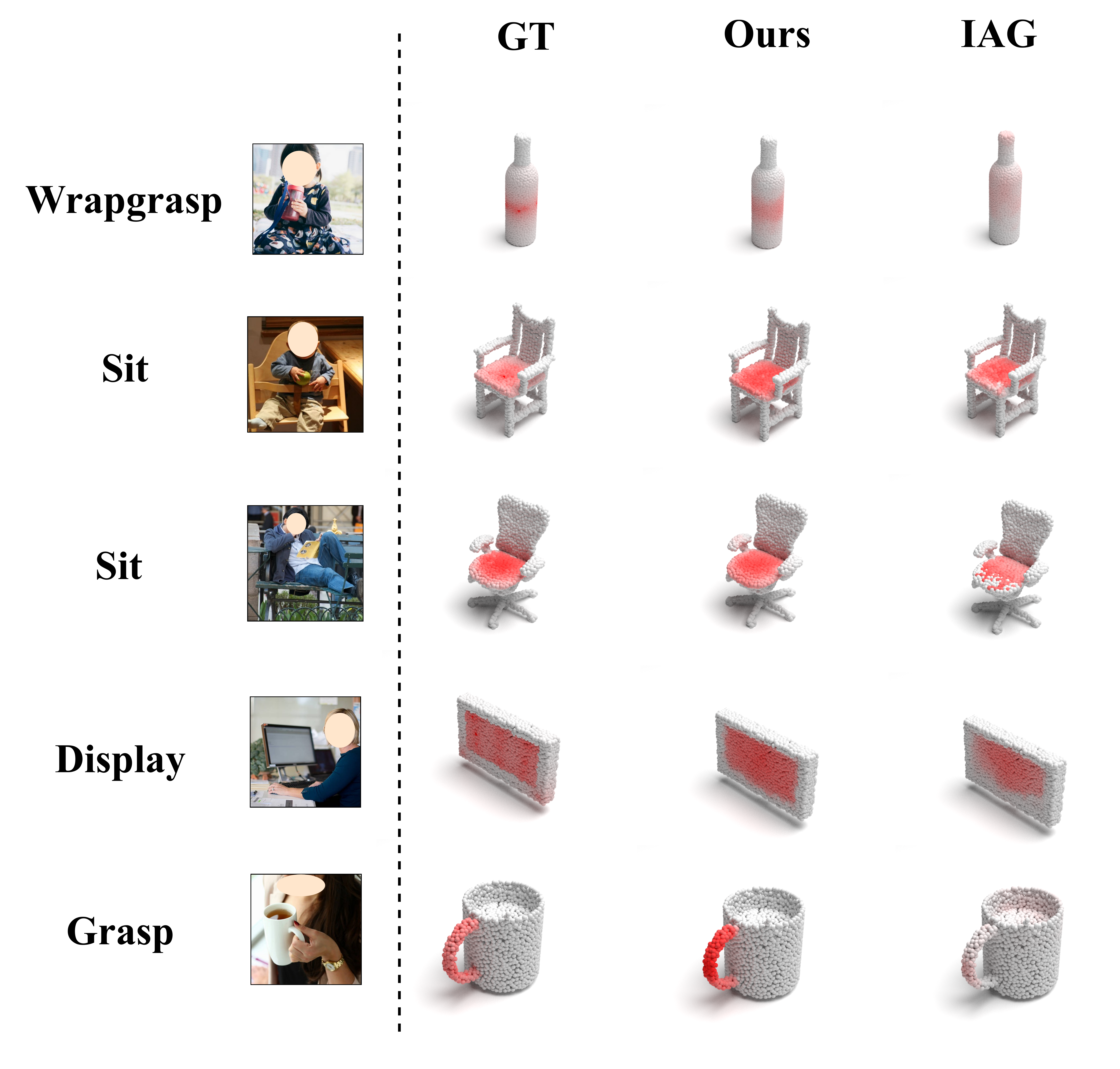}
    \caption{The visualization results of our method with IAG and GT.}
    \label{more_visualize1}
\end{figure*}

\begin{figure*}
    \centering
    \includegraphics[width=1\linewidth]{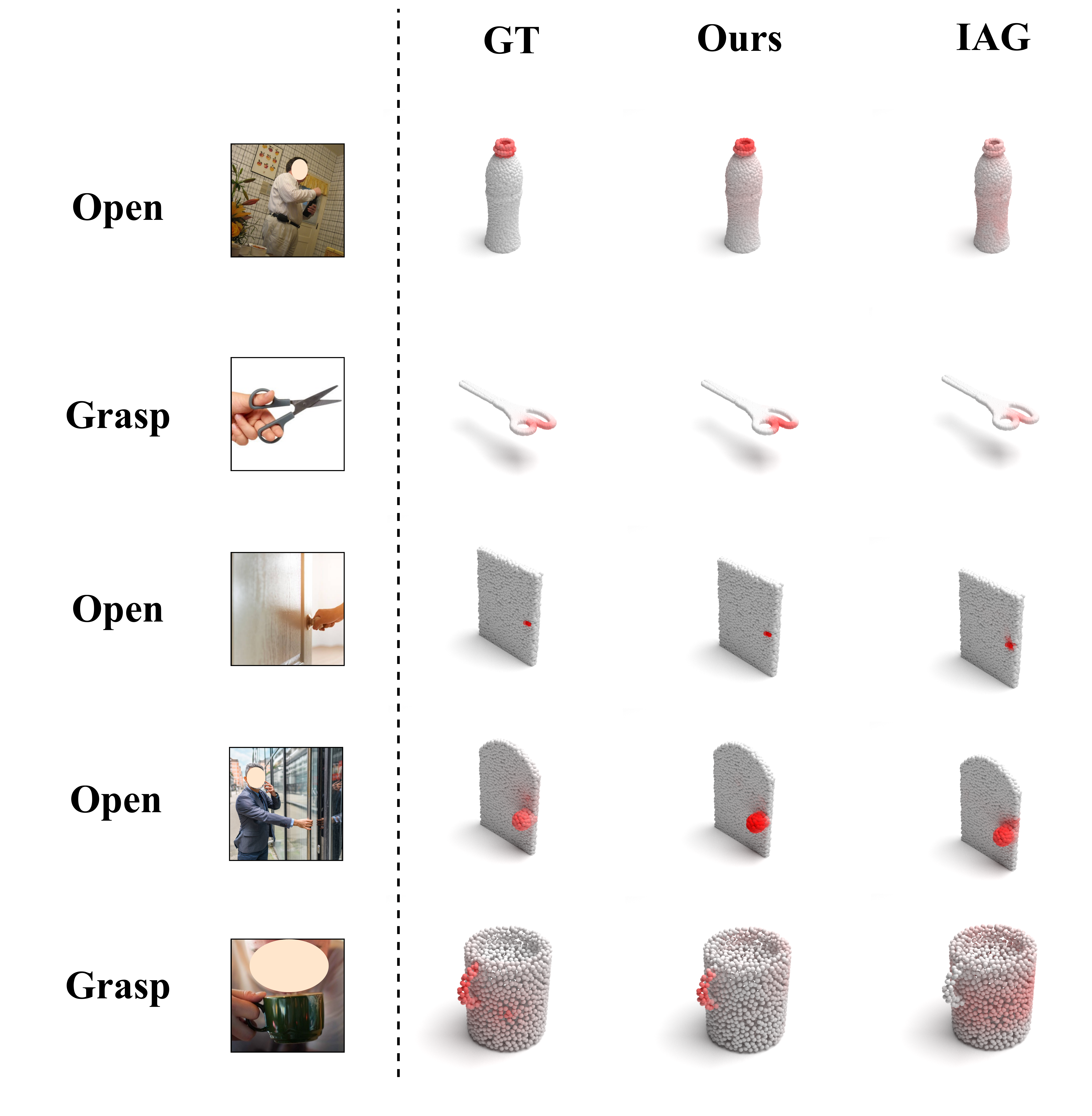}
    \caption{The visualization results of our method with IAG and GT.}
    \label{more_visualize2}
\end{figure*}

In the main text, we have provided several visualization results to demonstrate the effectiveness of our algorithm. In this section, we present additional visualization results to showcase the superior performance of our method, as shown in Fig. \ref{more_visualize1} and \ref{more_visualize2}. From these figures, it can be observed that our algorithm achieves satisfactory affordance grounding results across different affordances and various objects. Furthermore, the visualizations also demonstrate that our method successfully addresses the issues present in existing methods, such as ignoring multi-scale information and failing to incorporate the structural guidance of the point cloud for affordance grounding.

\subsubsection{Different instances}
\label{different instances}
\begin{figure}
    \centering
    \includegraphics[width=1\linewidth]{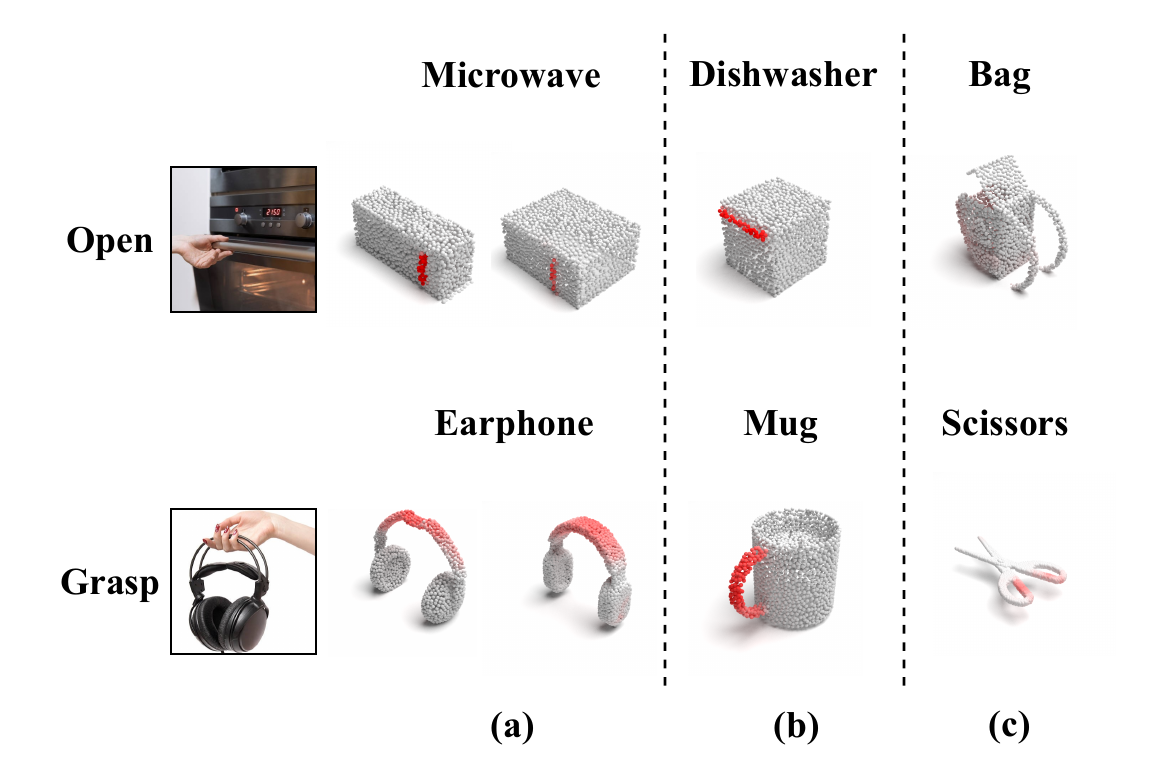}
    \caption{Comparison of the Same Image Across Multiple Point Clouds. (a) Identical object category. (b) Distinct object categories exhibiting similar geometrical features. (c) Different object categories with varied geometrical structures.}
    \label{sameimage}
\end{figure}

Unlike existing methods using a complex alignment module, we propose a simple yet effective cross-modal fusion module. Since we do not utilize a complex 2D-3D alignment module, we might still achieve satisfactory results when the the object in image and the point cloud share similar affordances. To investigate it, we conduct experiments by pairing the same image with: 1) objects of the same category, 2) objects of different categories but with similar geometric structures, and 3) different object categories with different geometric structures. The results are plotted in Fig. \ref{sameimage}. We observe that as the geometric discrepancy increases, the grounding performance gradually decreases. However, even for objects with vastly different geometric structures, our method can still achieve satisfactory results.

\subsubsection{Different affordances}
\label{different affordances}
\begin{figure}
    \centering
    \includegraphics[width=1\linewidth]{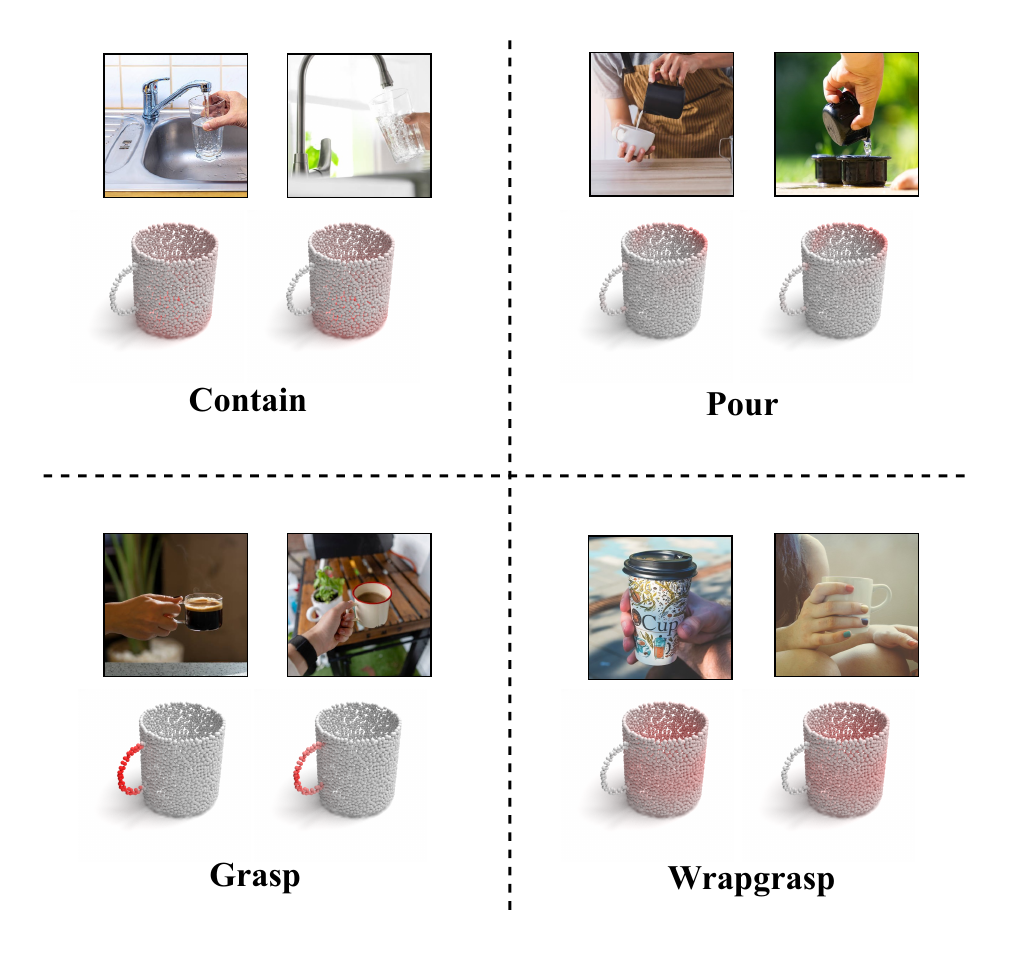}
    \caption{Comparison of the Identical Point Cloud among Multiple Images with Different Affordance Categories.}
    \label{same_object}
    \vspace{-1em}
\end{figure}

Our method first extract the context-aware affordance information of an image and utilize it to guide 3D affordance grounding. To verify the effectiveness of our image context learning, we pair the same point cloud with images that reflect different affordances, and we utilize our method to generate the grounding results. As shown in Fig. \ref{same_object}, we can observe that our method successfully grounds the affordance depicted in the image onto the 3D point cloud. Therefore, the 2D Context-Aware Affordance Extractor of our method is effective.





\end{document}